# The structure of verbal sequences analyzed with unsupervised learning techniques

Catherine Recanati, Nicoleta Rogovschi, Younes Bennani

LIPN - UMR 7030
CNRS - Université Paris 13
F-93430 Villetaneuse, France
Firstname.Secondname@lipn.univ-paris13.fr

**Abstract**
Data mining allows the exploration of sequences of phenomena, whereas one usually tends to focus on isolated phenomena or on the relation between two phenomena. It offers invaluable tools for theoretical analyses and exploration of the structure of sentences, texts, dialogues, and speech. We report here the results of an attempt at using it for inspecting sequences of verbs from French accounts of road accidents. This analysis comes from an original approach of unsupervised training allowing the discovery of the structure of sequential data. The entries of the analyzer were only made of the verbs appearing in the sentences. It provided a classification of the links between two successive verbs into four distinct clusters, allowing thus text segmentation. We give here an interpretation of these clusters by comparing the statistical distribution of independent semantic annotations.

## Introduction

Many studies emphasize the importance of tense in the narrative structure of texts (see (Vuillaume, 1990) for French narratives). Much has been written about the contrast between French *passé simple* and *imparfait*. [See (Hopper, 1979) for a good description of the opposition of these two tenses with respect to narrative structure, focus, and aspect]. Concerning aspect, it has been shown that one could not carry out the analysis of the succession of events in time using only tenses without referring to aspect (see e.g. Kamp, Vet or Vlach in (Martin and Nef, 1981), (Kamp and Rohrer, 1983), (Vet, 1994), (Gosselin, 1996), etc.). Aspect is initially defined as the domain of temporal organization of situations. It determines their localization in time but also introduces points of view. Therefore, there are links between aspect and other semantic domains, such as intentions or causality. Although all words of a sentence may contribute to aspectual meaning, verbs play a crucial role and lexical aspectual categories of verbs do exist. In this work, we shall attempt to detect regularities in sequences of verbs (within sentences) by focusing only on their tense and lexical aspectual category.

The French texts we analyze are short accounts of road accidents intended for the insurers. Their main purpose is to describe an accident, its causes, and to identify those who are responsible. The verbs are reduced here to pairs (*category*, *tense*), where *category* is one of the four lexical aspectual categories of a verb, and *tense* its grammatical tense.

We sought here to isolate typical sequences of such pairs on the assumption that they are meaningful, at the very least for the type of account considered. Let us add that the mathematical tools used here make it possible to check the statistical validity of the categories obtained, and that our semantic validation has been carried out with annotations unused by the training process.

## Advantages of our formal approach

One of the interests of unsupervised training is to allow the discovery of initially unknown categories. In this framework, the connectionist Self Organizing Maps (SOM) of Kohonen (Kohonen, 1995) provide an efficient categorization with simultaneous visualization of the results. This visualization is given by the topological map of the data (two similar data are close on the map) providing at the same time an "intelligent" coding of the data in the form of prototypes. Since these prototypes are of same nature as the data, they are interpretable, and the map thus provides a summary of the data.

From this coding, we took the Hidden Markov Models (HMM) to model the dynamics of the sequences of data (here, the verbs within a sentence). The HMM (Rabiner and Juang, 1986) are the best approach to treat sequences of variable length and to capture their dynamics. This is the reason why these models have been widely used in the field of voice recognition and are particularly well adapted to our objective.

To validate our hybrid approach, we used biological gene sequences and the textual data whose interpretation is given here. For technical details see (Rogovschi and al., 2007).

## Data encoding

We encoded a hundred or so texts containing 700 occurrences of verbs. In these texts, we considered all the sequences of at least two verbs delimited by the end of a sentence. To cope with the paucity of the data, we used re-sampling techniques based on sliding windows, which increase the redundancy (redundancy ensures a better classification of the data). For coding, the four aspectual categories of verbs introduced in (Vendler, 1967), namely state, activity, accomplishment and achievement, were combined with the tenses of the verb. The indexing is based on some of the ideas presented in (Recanati C. and Recanati F., 1999). Table 1 below briefly summarizes our understanding of these four categories.

These accounts of road accidents mostly use the *imparfait* (24%) and the *passé composé* (34%), with a few sentences in the present. In addition, there are also some (rare) occurrences of *passé simple* and of *plus-que-parfait*. There are however a significant number of present participles (11%) and infinitives (20%). We decided to retain all these "tenses" and carried out the training by using nine codes[1].

| STATE | ACTIVITY |
|---|---|
| homogeneous, often durative, habitual or dispositional | (a priori unbounded) durative process, macro-homogeneous |
| *be / expect / know* | *drive / run / zigzag* |
| **ACCOMPLISHMENT** | **ACHIEVEMENT** |
| durative process ending by a culmination point | change of state (near punctual) |
| *go to / park* | *reach / hit / break* |

Table 1: The four aspectual categories of verbs

**Example** « Le véhicule B *circulait* sur la voie de gauche des véhicules *allant* à gauche (marquage au sol par des flèches). Celui-ci *s'est rabattu* sur mon véhicule, me *heurtant* à l'arrière. Il *a accroché* mon pare-choc et m'*a entraîné* vers le mur amovible du pont de Gennevilliers que j'*ai percuté* violemment. » This description will be first reduced to the sequences of verbs: (circulait, allant) / (s'est rabattu, heurtant)/ (a accroché, a entraîné, ai percuté) – which are then numerically encoded as (tense, category) pairs: (act., IM) (acc., pp) / (acc., PC) (ach., pp) / (ach., PC) (acc., PC) (ach., PC).

## First results

The first results are the percentages of tenses and aspectual categories (independently of sequences of verbs). The verbs of state account for 24% of the corpus, of activity only 10%, verbs of accomplishment 34% and of achievement 32%. The percentages of tenses by category given graphically in Figure 1 confirm the interest of our pairings (*tense, category*).

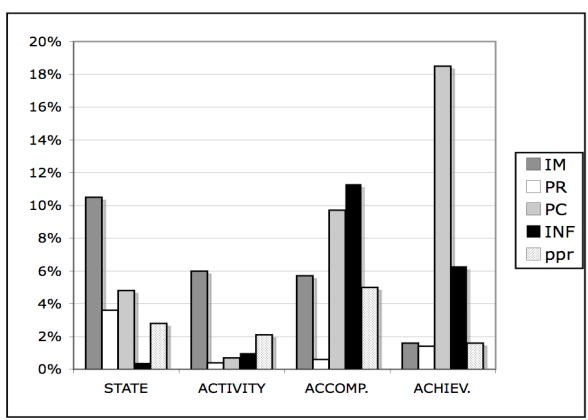

Figure 1: Distribution of main tenses by category

The nature of the aspectual categories, the aspectual specialization of grammatical tenses, and the typical structure of these accounts explain these percentages rather naturally.

---

[1] IM = *imparfait*, PR = *présent*, PC = *passé composé*, PS = *passé simple*, PQP = *plus-que-parfait*, inf = *infinitif*, ppr = *participe présent*, pp = *participe passé* and pps = *participe passé surcomposé*.

**Verbs of states (24%).** More than 70% are in the *imparfait*, the *present*, and the *participe présent*. This is not surprising since states are homogeneous and often durative, or characterize a disposition (habitual, generic). We find also a significant proportion of *passé composé* in connection with verbs like "want" or "can" ("j'ai voulu freiner", "je n'ai pu éviter" - "I wanted to slow down", "I could not avoid"). The small proportion of present arises from the fact that the account is in the past and the historic present is a literary style not appropriate for this kind of account.

**Verbs of activities (10%).** Similarly, the activities indicating unbounded processes, verbs of activities are distributed quite naturally with more than 79% in the *imparfait* tense and in the form of present participles. That 10% of the verbs are in the infinitive can be easily explained by the fact that activities are processes which may have a beginning, and which can thus be the complement of verbs like "start", "want", or simply can be introduced to mention a future goal of the driver (with the preposition "*pour*" + infinitive in French).

**Accomplishments (34%) and Achievements (32%).** Contrary to the two preceding categories, the telic character of these verbs explains their frequency in the *passé composé*. Achievements are mostly in the *passé composé* because denoting mainly a change of state, they are punctual (or of short time span), and take the *imparfait* tense only infrequently. In contrast, accomplishments often occur in the *imparfait* and as present participles, because they stress rather the process than its end - which brings them very close to activities. The global importance of these two categories (66%) has some connection with text typology, because an account of an accident implies a description of the sequence of the successive events that caused it.

**Aspectual specialization of tenses.** For (Smith, 1991), there are three points of view in the aspectual system of French. A perfective viewpoint shows a situation as being closed. It is expressed by the *passé composé* and the *passé simple*. The imperfective and neutral viewpoints present on the contrary open situations. A neutral viewpoint is expressed by the present. An imperfective viewpoint is expressed by the *imparfait*, or by the locution "*en train de*" (it would nevertheless be a mistake to assign systematically an imperfective aspect to all *imparfait* (Ducrot, 1979)). However, the opposition perfective/imperfective will be mainly embodied in these texts by the opposition *imparfait/passé composé*.

**A typical structure.** A description of an accident generally starts with one or two sentences describing the circumstances before the accident. This first part of the text is thus in the *imparfait*, and contains many present participles. In addition, there are also a few verbs in the present tense and many infinitives introduced by "*pour*", or occurring as complements of other verbs ("je m'apprêtais à tourner", "le feu venait de passer au rouge": "I was on the point of turning", "The stop light had just turned red"). This first part is mainly circumstantial and contains a majority of verbs of states, but also some of activities and of accomplishments. It is characterized by an overall imperfective point of view, and the account is in background.

The next part of the text contains a description of the accident that mentions the succession of events leading to the accident and finishes with the crucial moment of the impact. This indispensable part of the text uses mostly verbs of accomplishment and achievement, generally in the *passé composé*. It is characterized by a perfective mode, but since the goal is to indicate the responsibilities of the various actors, one still finds here many present participles and infinitive constructions connecting several verbs ("J'ai voulu freiner pour l'éviter" : "I wanted to slow down to avoid it"). At the end of the account, one occasionally finds a section with comments on the accident. That section often contains an inventory of the damage. This third part is often short and less easy to characterize.

Note that the structure we have indicated is only typical. The three parts can come in a different order, or some of them may be missing all together.

## Categorization of verbal sequences

Our unsupervised approach provided a classification of the sequences of two successive verbs (within the same sentence) in four groups (or clusters). The profiles of the resulting transitions are represented on maps (SOM) incorporating a notion of proximity. The matrix of the distances between the profiles of transitions provides a distribution of these transitions. That the profiles of the transitions fall into four distinct clusters accords with the Davies and Bouldin quality standard of unsupervised classification (Davies and Bouldin, 1979). In addition, it should be pointed out that since this number of four is small, it is to a certain extent a confirmation of the fruitfulness of our tense/lexical category pairing.

## Semantic interpretation

To provide the interpretation of the clusters, we carried out a certain number of semantic annotations. Thus, to account for the typical structure of these texts, we indexed all the verbs with a number indicating the part of the text in which they occurred (1-*circumstance*, 2-*accident* and 3-*comment*). We have also marked some of the verbs with the attributes *foreground* or *background* to indicate the type of contribution they make to the narrative structure. In order to be able to detect possible causal chains of verbs leading to the accident, we marked some verbs with the attributes *causal* and *impact* (*causal* when the verb was describing a direct cause of the accident, and *impact* when indicating the impact itself).

We also marked the verbs of action according to the agent responsible for the action (*A* for the driver and author of the text, *B* for the second driver involved in the accident, and *C* for a third person). We also noted the presence of negation, and the description of objectives or possible worlds that did not occur (attribute *inertia* for goals and alternatives).

The marking of negation was not very discriminating, and that of agent not very helpful. Table 2 summarizes the main results that we obtained by doing statistical comparisons of our semantic marks within these four clusters.

| **Cluster (IC)** *Impact and Comments* Very strong causality, foregrounding, frequent impact, goals and alternatives | **Cluster (AA)** *Actions leading to the Accident* Strong causality, neutral foreground/background, few impacts, many goals and alternatives |
|---|---|
| **Cluster (CA)** *Circumstances or Appearance of an incident* Little causality, background, little impact, no goal or alternative | **Cluster (C)** *Circumstances* No causality, background, no impact, many goals and alternatives |

Table 2: Summary of statistical interpretations

### Cluster (C) of Circumstances

The cluster (C) makes a clear distinction between the first and the second verb of a sequence. The first verb is 93% in the *imparfait*, but only 7% in the present, while the second is 63% in the infinitive and 30% in the present participle. From the point of view of aspectual categories, the first verb is a verb of state 56% of the time, and the second a verb of accomplishment 63% of the time. (The other categories divide themselves in a regular way between 12% and 16%). One can described the pairs of this group as made up of a verb of state (or activity) in the *imparfait*, followed by a verb of accomplishment in the infinitive or in the present participle. The analysis of the prototypes provides a finer synthesis (see Table 3).

This group favors states and activities to the detriment of achievements - accomplishments on the other hand are overwhelmingly present in the second verbal occurrence.

The cluster (C) is the one where the *inertia* attribute (indicating a goal or an alternative world) is the most frequent. This is explained by the many accomplishments introduced by the French preposition "pour" ("je reculais pour repartir", "je sortais du parking pour me diriger" - "I backed up to set out again", "I left the carpark to find my way"), or by the auxiliaries in the *imparfait* introducing an infinitive indicating the intentions of the driver ("j'allais tourner à gauche": "I was on the point of turning left"). This is one of the reasons why we called this cluster (C) "Circumstances". Another reason is that it contains a strong majority of verbs belonging to the first part of the account (63%), and very little to the second or the third. Moreover, this cluster hardly contains verbs indicating the causes of the accident or describing the impact. Actor A is represented the most there, and the account is in background.

### Cluster (CA) of Circumstances or of the Appearance of an incident

The second cluster (CA) is that of the circumstances or of the appearance of an incident. One notes here a great number of verbs of states (37.5%) and activities (17%), even more than in the preceding cluster and much higher than average. There are, on the other hand, an average number of achievements (29%), absent from the first verb but quite often present on the second. This distinguishes this cluster from the preceding one, where accomplishments played this role. Here, on the contrary, accomplishments are excluded from the second place, and are clearly under-represented (16.5%). One can synthesize the transitions of this cluster by saying that one generally has a state or an activity in the *imparfait* tense, followed by a state or an achievement in the *imparfait* or in the *passé composé*. (One thus connects two imperfective points of view, and sometimes, an imperfective point of view and a perfective one).

In this group 36% of verbs come from the first part of the account ("je circulais à 45 km/h environ dans une petite rue à sens unique où des voitures étaient garées des deux côtés": "I drove at approximately 45 km/h in a small one-way street where cars were parked on each side"), but also sequences finishing by an achievement in the *passé composé*, coming from the second part (34%, "je roulais rue Pasteur quand une voiture surgit de la gauche": "I drove on Pasteur street when a car emerged from the left"). This cluster contains also 25% of verbal sequences located between the two parts (providing thus half of such sequences). This is why we called it the cluster of circumstances or of the appearance of an incident. The account is here mainly in background. Actor A (or a third person C) are strongly indexed to the detriment of actor B. There are few allusions to the causes of the accident and to the impact, and few verbs in the foreground. The evocation of goals or alternative worlds is missing from this group (contrary to the preceding group).

### Cluster (AA) of Actions leading to the Accident

The third cluster (AA) clearly marks the account of the accident itself. It is characterized by the wealth of achievements, to the detriment of states and activities. It is in perfective mode but includes many infinitives. The detected prototypes are given in Table 3. These pairs lend themselves well to constructions of three verbs as in, "J'ai voulu m'engager pour laisser" ("I wanted to turn into to let"), or "n'ayant pas la possibilité de changer de voie et la route étant mouillée" ("not having the possibility of changing the lane and the road's being wet"). Fifty-six percents of the pairs come from the second part of the account. This is why, since this cluster also contains many accomplishments, we called it a cluster of actions leading to the accident. Nevertheless, as in cluster (C), the present participles and infinitives allow the expression of goals and alternative worlds ("désirant me rendre à", "commençant à tourner" - "wishing to go to", "starting to turn"), so 26% of the pairs come from the first part of the account. We noted however a rather strong proportion of agent A and B, with little of C, but also little foregrounding - the account being unmarked (neither in foreground, nor in background). Many verbs describe the causal chain, but relatively few mention the impact.

### Cluster (IC) of the Impact and of the Comments

The verbs of achievements (45%) appear here in a larger number than elsewhere (32% on average), to the detriment of activities (6.5%), and states (only 14.5%). This explains that this group supports the descriptive part of the accident (57%). Here also one observes an increase in infinitives and participles on the first verb to the detriment of the *imparfait* and of the present, and a large increase in *passé composé* on the second verb to the detriment of all tenses - except the present (8%, slightly higher than the average). Perhaps the occurrence of the present is related to the strong proportion of comments (29%), more important than elsewhere (18% on average). The mention of goals or alternatives is average. It is here on the other hand that foregrounding is most marked. There is an important number of references to actor B (the driver of the other vehicle) and it is here that one finds most verbs relating to the causes of the accident and the impact itself. Table 3 indicates only two typical elements in this group, which, although quite large, is more difficult to characterize. The analysis in terms of viewpoints shows nevertheless that the sequence generally finishes with a perfective viewpoint.

### Typical pairs

Information concerning the Markov chains (transferred back to the topological map) enabled us to reduce the number of typical pairs (verb1, verb2) by pruning. A synthesis is given in Table 3.

| Types | verb 1 | verb 2 |
|---|---|---|
| C    1 | state or act. IM | state or act., ppr |
|       2 | state IM (or PR) | acc., INF |
|       3 | act. or ach. IM | acc. (or ach.) INF |
| CA   4 | state or act. IM | state (or ach.), IM |
|       5 | state or act. IM | state (or ach.), PC |
| AA   6 | acc.(or ach.) INF | acc.( or ach.) INF (or ppr) |
|       7 | ach.(or acc.) PC | acc.( or ach.) INF |
|       8 | state PC | ach. INF |
| IC   9 | ach.(or acc.) INF | ach. PC |
|      10 | ach. or state, PC | ach.(or acc) PC |

Table 3: Most typical pairs of the clusters

### Comments

This categorization distinguishes quite well states and activities (groups C, CA) from telic events (groups AA, CC). In a more interesting way, accomplishments are also distinguished from achievements, justifying the distinction in opposition to a single notion of telic events. It should also be noticed that the expression of goals or alternatives often accompanies the use of verbs in the present participle or infinitive - which explains the ratio displayed by groups C and AA. However, the lexical category used influences also this expression, because the second verb in these two groups is generally an accomplishment. Moreover, the groups C and CA (unmarked for this purpose) differ precisely in the type of event appearing in second place. In the same way, elements differentiating the groups AA and CC (which contain a majority of telic events) show that the group AA (which favors accomplishments), although conveying a perfective mode, is little marked for narrative foreground. This group is also less concerned with the causes of the accident than the group CC, and it makes little allusion to the impact. Goals and intentions would thus be expressed more easily by accomplishments than by achievements - which would carry more causality. Indeed, the group CC, which favors achievements, is more strongly marked for narrative foregrounding, the impact itself, and the causal chain of events, which directly caused it. It may be that for achievements and activities, the subject has a relation of power over the direct object (or over the oblique object). One can check its existence by using adverbs of manner (gently, with caution, etc). This explains perhaps also the strong attribution of responsibility to the agent with the use of verbs of achievements.

Although the perfective/imperfective contrast emerges rather neatly (clusters AA and CC vs. C and CA), surprisingly cluster (CA) hosts both imperfective sequences and *imparfait*/*passé composé* sequences. One of the explanations is that our training algorithm did not the linking of two sentences take into account, so that the succession of several sentences in the imperfect mode, and the typical structure of these accounts (such as we perceive it) could not be detected. One thus misses a significant part of our objective. However, the results obtained in focusing only on sentences are still interesting, since the three typical parts that we have distinguished fall differently onto the four clusters, although not in a uniform way. It should also be noted that this classification underlines the importance of infinitive turns and present participles, and the subtlety of their linking (cf. Table 3).

**Technical improvements**. One built the HMM here by moving a window of size 2: verbs are analyzed by taking into consideration the verbs that precede and follow them, but not the N-preceding verbs or the N-following. This is not important here, since our analysis is at the sentence level, (and in these accounts, a sentence contains rarely more than three verbs), but for an analysis of the entire structure of a text, we will need to add this. In addition, we would have liked to produce typical sequences of variable length instead of simple pairs. Thus, clusters AA and CC would have provided sequences of three verbs. (This follows from the prototypes of Table 3, and was noted on the segmentation of texts). This result could be obtained automatically (because the HMM provides the probabilities of transitions between two verbs), but we have not had enough time to implement this method.

## Conclusion

Our general project is to apply techniques of data mining to explore textual structures. For this purpose we developed an original approach of unsupervised learning allowing the detection of sequential structures. We applied it to the analysis of sequences of verbs in sentences coming from a corpus of accounts of road accidents. We obtained a classification of pairs of two successive verbs in four groups (or clusters). We succeeded in validating these groups satisfactorily, by basing our judgment on semantic annotations and the application of statistical analyses. This validates at the same time the power of the technique employed, and that of our coupling of the grammatical tense with the aspectual lexical category of a verb. However, this work is still in its early stages, and many points remain to be elucidated. We first regret not to have been able to compare our statistical analysis on cross tense/category uses with those of other types of texts (and in particular with that of simple accounts of incidents). It indeed remains to be determinded what the "typological" part of the isolated sequences is. Nor did we have time to exploit the probabilities of transitions given by the HMM; but this could be an interesting application of the techniques developed in this paper. The general method should also be improved so as to take into account the entire structure of a text (which is not the case here), and make room for sequences of length higher than 2.

## Acknowledgments

We warmly thank Steven Davis and Adeline Nazarenko for their help with respect to both substance and style.

## References

Davies, D.L. and Bouldin, D.W. (1979). A Cluster Separation Measure. In IEEE Transactions on Pattern Analysis and Machine Learning, 1(2).

Ducrot O. (1979). L'imparfait en français. In *Linguistische Berichte* n°60, (pp 1--23).

Gosselin L. (1996). Sémantique de la temporalité en français. Louvain-la-Neuve : Duculot.

Hopper J. (1979). Some observations on the typology of focus and aspect in narrative language. In Studies in Language 3.1 (pp. 37--64). Amsterdam: J. Benjamins.

Kamp H., Rohrer C. (1983). Tense in Texts. In Bauerle R., Schwarze C. et von Stechow A. (eds), Meaning, Use and Interpretation of Language, (pp. 250--269). Berlin : De Gruyter.

Kohonen T., (1995). Self-Organizing Map. Springer.

Martin R, Nef F. (eds) (1981). Le temps grammatical. In Langage n°64, Kamp H. (pp. 39--64), Vlach F. (pp. 65--79), Vet C. (pp. 109--124), Paris : Larousse.

Rabiner L.R., Juang B.H. (1986). An Introduction to Hidden Markov models. In IEEE ASSP Magazine, jan. 86, (pp. 4--16).

Recanati C. and Recanati F. (1999). La classification de Vendler revue et corrigée. Cahiers Chronos 4, La modalité sous tous ses aspects (pp. 167--184). Amsterdam/Atlanta, GA.

Rogovschi N., Bennani Y., Recanati C. (2007). Apprentissage neuro-markovien pour la classification non supervisée de données structurées en séquences. In Actes des 7èmes journées francophones Extraction et Gestion des Connaissances. Namur, Belgique.

Smith C. S. (1991). The parameter of aspect. Studies in Linguistics and Philosophy. Kluwer Academic publishers.

Vendler Z. (1967). Verbs and Times. In Linguistics in Philosophy, (pp. 9--121). Ithaca, New-York: Cornell University Press.

Vet C. (1994). Relations temporelles et progression thématique. In Études Cognitives 1, Sémantique des Catégories de l'aspect et du Temps, (pp. 131--149). Warszawa : académie des Sciences de Pologne.

Vuillaume M. (1990). Grammaire temporelle des récits. Paris : Minuit.